\documentclass[letterpaper]{article} 
\usepackage{aaai25}  
\usepackage{times}  
\usepackage{helvet}  
\usepackage{courier}  
\usepackage[hyphens]{url}  
\usepackage{graphicx} 
\usepackage{natbib}  
\usepackage{caption} 
\usepackage{algorithm}
\usepackage{algorithmic}

\usepackage{newfloat}
\usepackage{listings}
\DeclareCaptionStyle{ruled}{labelfont=normalfont,labelsep=colon,strut=off} 
\floatstyle{ruled}
\newfloat{listing}{tb}{lst}{}
\floatname{listing}{Listing}
\usepackage{xcolor}
\usepackage{makecell}
\usepackage{pifont}\usepackage{tcolorbox}
 \usepackage{booktabs}
 \usepackage{xspace}
 \usepackage{outlines}
 \usepackage{subfigure}
 \usepackage{amsmath}
 \usepackage{multirow}
 \usepackage{enumitem}
 \newcommand{\ourmethod}{{HUGE}\xspace}
 \newcommand{\ourhetero}{{HALO}\xspace}
\usepackage{amsthm} 
\newtheorem{theorem}{Theorem}

\theoremstyle{definition}

\title{A Label-Free Heterophily-Guided Approach \\for Unsupervised Graph Fraud Detection}
\author {
    Junjun Pan\textsuperscript{\rm 1},
    Yixin Liu\textsuperscript{\rm 1},
    Xin Zheng\textsuperscript{\rm 1},
    Yizhen Zheng\textsuperscript{\rm 2},\\
    Alan Wee-Chung Liew\textsuperscript{\rm 1},
    Fuyi Li\textsuperscript{\rm 3},
    Shirui Pan\textsuperscript{\rm 1}\thanks{Corresponding  Author}
}
\affiliations {
    \textsuperscript{\rm 1}School of Information and Communication Technology, Griffith University, Queensland, Australia \\
    \textsuperscript{\rm 2}Faculty of Information Technology, Monash University, Melbourne, Australia\\
    \textsuperscript{\rm 3}Faculty of Health and Medical Sciences, The University of Adelaide, South Australia, Australia\\
    \{junjun.pan, yixin.liu, xin.zheng, a.liew, s.pan\}@griffith.edu.au, yizhen.zheng1@monash.edu, 
    fuyi.li@adelaide.edu.au
}

\usepackage{appendix}

\begin{document}

\maketitle

\begin{abstract}

Graph fraud detection (GFD) has rapidly advanced in protecting online services by identifying malicious fraudsters. Recent supervised GFD research highlights that heterophilic connections between fraudsters and users can greatly impact detection performance, since fraudsters tend to camouflage themselves by building more connections to benign users. Despite the promising performance of supervised GFD methods, the reliance on labels limits their applications to unsupervised scenarios; Additionally, accurately capturing complex and diverse heterophily patterns without labels poses a further challenge. To fill the gap, we propose a \textbf{H}eterophily-guided \textbf{U}nsupervised \textbf{G}raph fraud d\textbf{E}tection approach (HUGE) for unsupervised GFD, which contains two essential components: a heterophily estimation module and an alignment-based fraud detection module. In the heterophily estimation module, we design a novel label-free heterophily metric called HALO, which captures the critical graph properties for GFD, enabling its outstanding ability to estimate heterophily from node attributes. In the alignment-based fraud detection module, we develop a joint MLP-GNN architecture with ranking loss and asymmetric alignment loss. The ranking loss aligns the predicted fraud score with the relative order of HALO, providing an extra robustness guarantee by comparing heterophily among non-adjacent nodes. Moreover, the asymmetric alignment loss effectively utilizes structural information while alleviating the feature-smooth effects of GNNs. Extensive experiments on 6 datasets demonstrate that HUGE significantly outperforms competitors, showcasing its effectiveness and robustness. The source code of HUGE is at \url{https://github.com/CampanulaBells/HUGE-GAD}.

\end{abstract}

%

\section{Introduction}


With the advancement of information technology, graph-structured data have become ubiquitous in online services such as e-commerce~\cite{motie2023financial, zheng2022rethinking, zheng2024gnnevaluator,zheng2022unifying, wang2024contrastive,zhengonline} and social media~\cite{deng2022markov}. However, this widespread usage has led to a significant increase in various malicious activities, including fraud, spam, and fake reviews, making detecting such activities crucial. To address this issue, graph fraud detection (GFD) is an emerging research topic that aims to identify these suspicious activities. By analyzing both structural patterns and attribute contexts within networks, GFD ensures the integrity of online applications~\cite{motie2023financial,deng2022markov,yu2022graph}. 

\begin{figure}[!t]
\centering     
\subfigure[Heterophily-based supervised GFD method.]{\label{fig:a}\includegraphics[width=0.9\linewidth]{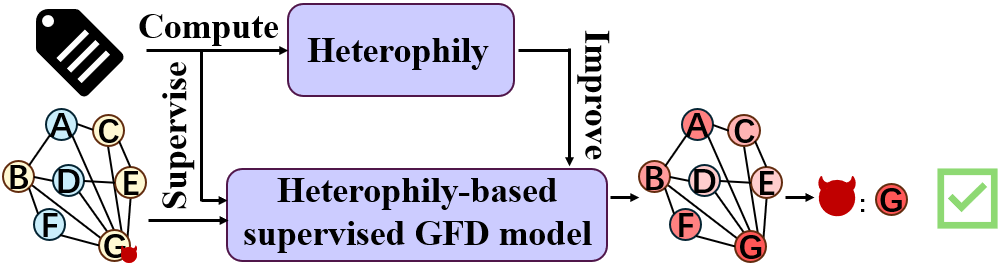}}
\subfigure[Unsupervised GAD method.]{\label{fig:b}\includegraphics[width=0.9\linewidth]{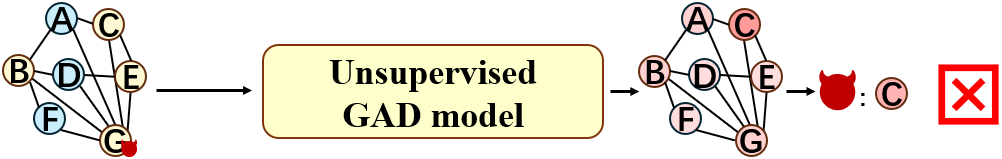}}
\subfigure[Our heterophily-guided unsupervised GFD method.]{\label{fig:c}\includegraphics[width=0.9\linewidth]{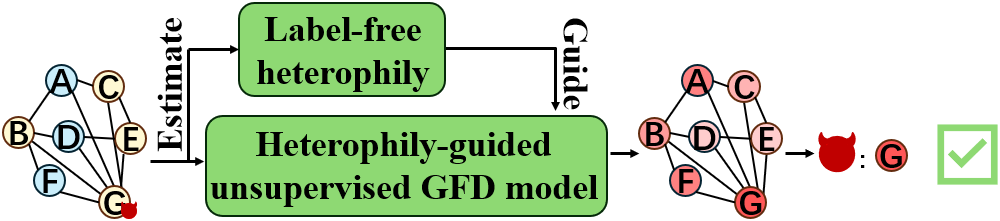}}
\caption{Concept maps of existing GFD/GAD methods and our proposed unsupervised GFD method \ourmethod. }
\label{fig:intro}
\end{figure}

Recently, supervised GFD methods based on graph neural networks (GNNs) have shown significant progress and attached considerable research interest~\cite{dou2020enhancing,gao2023addressing,li2022cyclic}. One of the mainstream research focuses is on addressing the graph \textbf{heterophily} introduced by fraudsters~\cite{gao2023addressing, zheng2022graph}, i.e., fraudsters tend to establish connections with benign users to avoid being detected for concealing their suspicious activities. In this way, conventional GNN models struggle to differentiate between the representations of fraudsters and benign users due to the smooth effect of message passing~\cite{cai2020note}, resulting in sub-optimal GFD performance. In light of this, various strategies have been proposed to mitigate the impact of heterophily, such as dropping edges~\cite{gao2023addressing}, or using spectral GNNs~\cite{tang2022rethinking}. 
Despite the promising performance of these methods, they rely heavily on annotated labels of both fraudsters and benign users for modeling the heterophily property. However, in real-world scenarios, obtaining annotated fraudsters typically requires significant human effort, making supervised GFD methods impractical when labels are unavailable. Therefore, developing \textit{unsupervised GFD methods} to avoid label reliance can be a promising research direction.

Recently, several studies have attempted to address the challenge of unsupervised GFD under graph heterophily by leveraging graph anomaly detection (GAD). GAD can be regarded as a superset of GFD, as it not only seeks to identify fraudsters but also any rare events or patterns within graph data~\cite{ma2021comprehensive}, thereby making it a viable solution for unsupervised GFD. 
However, these methods have primarily demonstrated promising results on artificial datasets within the general GAD context. The artificial anomalous patterns (e.g., dense subgraph and feature outliers) are relatively straightforward to identify~\cite{ding2019deep,gu2024three}. In contrast, in real-world fraud detection, fraudsters often introduce more heterophilic connections with benign users to conceal their activities. To address these challenges, recent works~\cite{he2024ada, qiao2024truncated} have sought to iteratively reduce heterophily from graph structure by estimating pseudo-labels. Nevertheless, the iterative refinement and multiple-round pseudo-label computation limit their time and memory efficiency. Moreover, estimating pseudo-labels using GNNs may still suffer from feature-smoothing effects, potentially misguiding the subsequent iterations. 
The general comparison between existing solutions and our proposed method is demonstrated in Figure~\ref{fig:intro}.\looseness-2

In light of the insufficiency of existing solutions for the unsupervised GFD problem under graph heterophily,  
a natural research question arises: 

\textit{Can we explicitly measure heterophily without annotated labels and utilize it to guide the learning of the GFD model?}

To answer this question, we identify two challenging aspects for effective unsupervised GFD:

\noindent\textit{\textbf{Challenge 1} - Precise unsupervised heterophily estimation: } 
the calculation of heterophily often relies on labels~\cite{zheng2022graph} that are unavailable in unsupervised GFD scenarios. While several attribute-level distance metrics have the potential to indicate heterophily, they hardly satisfy some crucial properties for effective GFD (e.g. Euclidean distance is unbounded, cosine distance is scale insensitive, etc.), limiting their applicability in unsupervised GFD. 

\noindent\textit{\textbf{Challenge 2} - Enabling effective heterophily guidance in GFD: } even though a proper unsupervised heterophily metric can be established, how to effectively use them for subsequent GAD remains a challenging task. 
Since the heterophily is estimated from attributes, it provides not only clues about fraudsters but also noise. Therefore, simply following previous supervised GFD methods may lead to sub-optimal results for unsupervised GFD. 

To address the above challenges, in this paper, we proposed a novel \textbf{H}eterophily-guided \textbf{U}nsupervised \textbf{G}raph fraud d\textbf{E}tection method (\ourmethod for short), which contains two essential components: a heterophily estimation module and an alignment-based fraud detection module, which provides a powerful solution for unsupervised GFD that overcomes previous challenges. Specifically, we analyze the desired properties that label-free heterophily measures should possess for GFD. Based on these, we design a novel metric called HALO to address \textbf{Challenge 1}, which captures the critical graph properties for GFD, enabling its outstanding ability to estimate heterophily from node attributes. 
In the alignment-based fraud detection module, we develop a joint MLP-GNN architecture with ranking loss and asymmetric alignment loss to address \textbf{Challenge 2}. The ranking loss aligns the order 
 of predicted fraud scores with that of \ourhetero to guide the model training process. Utilizing the rank of heterophily as guidance not only improves the robustness by mitigating irrelevant information, but also introduces extra global heterophily constraints by comparing heterophily among non-adjacent nodes. 
Moreover, the asymmetric alignment loss effectively extracts structural information from GNN to MLP by aligning the neighbor inconsistency distributions, alleviating the feature-smooth effect introduced by message passing. 
For evaluation, we conduct extensive experiments to demonstrate the superiority of \ourmethod over state-of-the-art methods on six real-world GFD datasets. In summary, our contributions are three-fold:

\begin{itemize} [noitemsep,leftmargin=*,topsep=1.5pt]
    \item We derive a new label-free heterophily metric termed HALO to accurately estimate heterophily by capturing the critical graph properties for unsupervised GFD. 
    \item We develop a novel unsupervised GFD framework, \textbf{H}eterophily-guided \textbf{U}nsupervised \textbf{G}raph fraud d\textbf{E}tection (\ourmethod). We first estimate heterophily from attributes with \ourhetero, then utilize a joint MLP-GNN architecture with ranking loss and asymmetric alignment loss to guarantee robustness and effectiveness. Finally, the local inconsistency scores of MLP embeddings are used as fraud scores to spot fraudsters. 
    \item We conduct extensive experiments to demonstrate the superior performance of \ourmethod over the state-of-the-art methods on six real-world GFD datasets under unsupervised learning scenarios.
\end{itemize}

\begin{table*}[t!]
\centering

\setlength{\tabcolsep}{1mm}
\fontsize{9pt}{9pt}\selectfont
\begin{tabular}{c | c | c c c c} 
 \toprule
 Metric & Definition & Boundedness & Minimal Agreement  & Monotonicity &  Equal Attribute Tolerance \\
 \midrule
 Euclidean Distance & $\|\mathbf{x}_{i}-\mathbf{x}_{j}\|$ & \ding{56} & \ding{52} & \ding{52} & \ding{52}\\
 Cosine Distance & $-\frac{{\mathbf{x}_{i}}^T\cdot{\mathbf{x}_{j}}}{\|\mathbf{x}_{i}\|\|\mathbf{x}_{j}\|}$ & \ding{52} &  \ding{56}  &  \ding{56}  &  \ding{56} \\
 Attr. Het. Rate  \cite{yang2021diverse} & $\frac{\sum^{d}_{k=1} { \mathbf{1}[x_{i,k}\neq x_{j,k}]} }{d}$ & \ding{52} &  \ding{52}  & \ding{56} &  \ding{56} \\
  \midrule
 \textbf{\ourhetero (ours)} & Eq. (\ref{equation:Hedge}) & \ding{52} & \ding{52} & $\text{\ding{52}}^{\star}$  & \ding{52} \\
 \bottomrule
\end{tabular}
\caption{Comparison of unsupervised edge-level heterophily metrics. \ding{52}* denotes that the metric satisfies the property under relaxed constraints. Vectors $\mathbf{x}_{i}$, $\mathbf{x}_{j} $ denote attributes of nodes $v_i$, $v_j$ respectively. Attr. Het. Rate refers to attribute heterophily rate. The indicator function $\mathbf{1}[x_{i,k}\neq x_{j,k}]$ returns 1 when the condition $x_{i,k}\neq x_{j,k}$ is satisfied, and 0 otherwise.}

\label{table:metrics}
\end{table*}

\section{Related Work}

\noindent \textbf{Graph Fraud Detection (GFD)} aims to identify fraudulent activities from graph-structured real-world systems, including financial fraud~\cite{motie2023financial}, spamming~\cite{deng2022markov}, and fake reviews~\cite{yu2022graph}. Various techniques have been utilized to detect the fraudsters, including attention~\cite{liu2020alleviating}, sampling~\cite{liu2021pick, liu2021pick} and multi-view learning~\cite{zhong2020financial}. 
Although these methods have achieved promising results, they suffer from heterophily-caused problem~\cite{dou2020enhancing}, i.e., fraudster nodes tend to build heterophilic connections with benign user nodes to make them indistinguishable from the majority.
Recent studies have developed many strategies to mitigate the negative impact of heterophily with node labels.
For example, GHRN~\cite{gao2023addressing} reduce heterophily by pruning the graph, while BWGNN~\cite{tang2022rethinking} leverage spectral GNNs to better capture high-frequency features associated with heterophily. GDN~\cite{gao2023alleviating}, GAGA~\cite{wang2023label}, and PMP~\cite{zhuo2024partitioning} overcome the feature-smoothing effect by crafting advanced encoders that sharpen feature separation, while ConsisGAD~\cite{chen2024consistency} directly utilizes annotated labels to create more effective graph augmentation method. 
Despite their promising results, the reliance on labels in existing methods restricts their applicability in unsupervised scenarios. Therefore, in this work, we aim to address the pressing need for developing unsupervised GFD methods.

\noindent \textbf{Graph Anomaly Detection (GAD)} is a broader concept than GFD, aiming to identify not only fraudsters but also any rare and unusual patterns that significantly deviate from the majority in graph data~\cite{ding2019deep, zheng2021heterogeneous,wang2024unifying,liu2024arc,cai2024lgfgad,cai2022plad,zhang2024deep,liu2024towards}. Therefore, GAD techniques can be directly applied to GFD, especially in unsupervised learning scenarios~\cite{li2024noise,liu2024self}. Given the broad scope of GAD and the difficulty in obtaining real-world anomalies, many unsupervised GAD methods have been designed and evaluated on several datasets with artificially injected anomalies~\cite{ding2019deep, liu2021anomaly, jin2021anemone, zheng2021generative, duan2023graph, duan2023arise, pan2023prem}.
Despite their decent performances, these methods rely on the strong homophily of the datasets with injected anomalies, which limits their applications under graph heterophily~\cite{zheng2022graph, zheng2023finding}. Recent studies explore this issue and suggest using estimated anomaly scores as pseudo-labels to mitigate the negative impacts of heterophily, e.g., dropping edges~\cite{he2024ada, qiao2024truncated} and adjusted message passing~\cite{chen2024boosting}. 
However, the estimated anomaly score might not be an effective indicator for modeling the heterophily under the homophily-based message-passing GNNs~\cite{zhu2022does}, leading to suboptimal GFD performance~\cite{GLOD}.
Hence, we introduce a novel heterophily measurement with desired properties as effective guidance for a powerful unsupervised GFD model.

More detailed related works are available in Appendix A.

\section{Preliminary}

\noindent\textbf{Notations.} Let $G=(\mathcal{V}, \mathcal{E}, \mathbf{X})$ represents an attributed graph, where $\mathcal{V}$ is the set of nodes and $\mathcal{E}$ is the set of edges. We denote the number of nodes and edges as $n$ and $m$, respectively. $\mathbf{X}\in\mathcal{R}^{n \times d}$ is the attribute matrix, with $d$ being the dimension of the attributes. The attribute vector of the $i$-th node $v_i$ is represented by its $i$-th row, i.e., $\mathbf{x}_i$. The graph structure can also be represented as adjacent matrix $\mathbf{A}\in\mathcal{R}^{n \times n}$, with $\mathbf{A}_{i,j}=1$ if there exists an edge between nodes $v_i$ and $v_j$, and $\mathbf{A}_{i,j}=0$ otherwise. We denote the set of neighbors of the $i$-th node as $\mathcal{N}(v_i)=\{{v}_j|\mathbf{A}_{i,j}=1\}$.

\noindent\textbf{Problem Definition.} In the context of graph fraud detection (GFD), each node $v_i \in \mathcal{V}$ has a label $y_i \in \{0, 1\}$, where $y_i=0$ indicates a benign node and $y_i=1$ represents a fraudulent node. An assumption is that the number of benign nodes is significantly greater than the number of fraudulent nodes. In unsupervised GFD scenarios, labels are unavailable during the training stage. The goal of a GFD method is to identify whether a node $v_i$ is suspicious by predicting a fraudulent score $s_i^{\text{fraud}}$, where a higher score indicates a greater likelihood that the node is fraudulent.

\begin{figure*}
\centering
\includegraphics[width=0.9\linewidth]{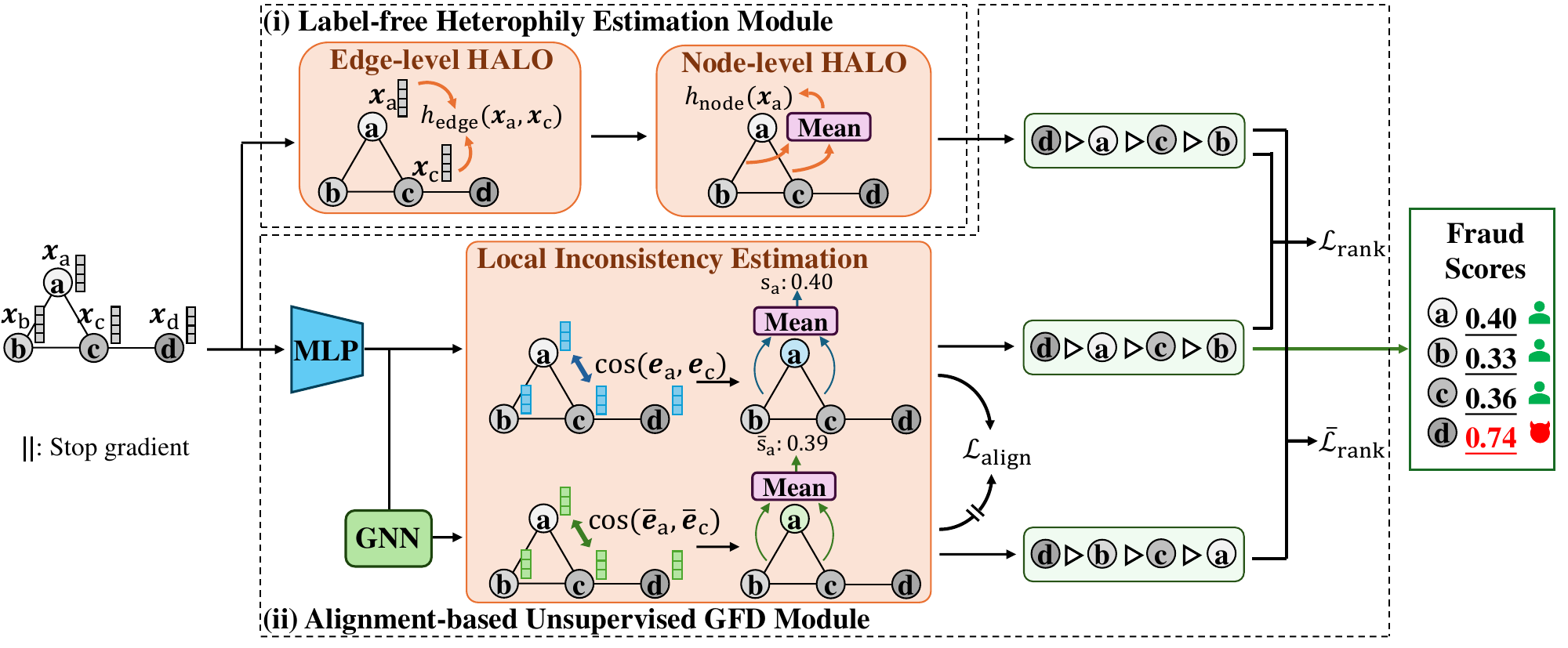}
\caption{Workflow of \ourmethod. Our \textit{label-free heterophily estimation module} estimates node heterophily using attributes. Then, in the \textit{alignment-based unsupervised GFD module}, a joint MLP-GNN architecture is trained through ranking and asymmetric alignment losses. 
The ranking loss ensures the predicted inconsistency score aligns with the heterophily order, while the asymmetric alignment loss matches the neighbor inconsistency distribution of the MLP encoder to that of the GNN encoder. 
In evaluation phase, the local inconsistency scores generated by the MLP encoder are used as the final fraud scores. }
\label{fig:architecture}
\end{figure*}
\section{Methodology}

In this section, we introduce the proposed unsupervised graph fraud detection (GFD) approach termed \textbf{H}eterophily-guided \textbf{U}nsupervised \textbf{G}raph fraud d\textbf{E}tection (\ourmethod). As illustrated in Figure~\ref{fig:architecture}, \ourmethod consists of two components at the highest level: label-free heterophily estimation module and alignment-based fraud detection module.
We first estimate the heterophily using a novel label-free heterophily measure termed \ourhetero. Then, the fraud detection model is trained following the guidance of the relative order of \ourhetero with a ranking loss and an asymmetric alignment loss, leading to an effective unsupervised GFD framework.
More details of the proposed \ourmethod are presented as follows.

\subsection{Label-free Heterophily Estimation}
GFD contends with a well-known challenge named the camouflage effects~\cite{dou2020enhancing}, wherein fraudsters tend to connect with normal nodes to hide their suspicious features. To address this issue, heterophilic edges serve as a crucial hint, revealing the feature of connections that fraudsters attempt to establish. 
However, under unsupervised application scenarios, it becomes significantly challenging to directly compute heterophily from attributes. Hence, in this work, we proposed to comprehensively and precisely estimate heterophily from node attributes.

In this subsection, we first discuss a set of universally desired properties for the estimation of heterophily in unsupervised GFD and provide a detailed review of the aforementioned insufficiency. We then introduce a new label-free heterophily measure named \ourhetero, which satisfies these properties and provides better guidance for detecting camouflaged fraudsters. 

\noindent \textbf{Desirable properties for heterophily measures.} Measuring the inconsistency between two nodes is the fundamental building block for modeling heterophily. 
While defining heterophily~\cite{zheng2022graph} from annotated labels is straightforward, extending its definition using node attributes is challenging, as attributes are continuous variables with multiple dimensions. Moreover, fraudsters may intentionally modify their attributes to appear normal~\cite{dou2020enhancing}, complicating heterophily estimation. 
Motivated by~\cite{platonov2024characterizing}, we summarize a set of desirable properties for edge-level label-free heterophily $h_{\text{edge}}$. 

\begin{itemize} [noitemsep,leftmargin=*,topsep=1.5pt] 
    \item \textbf{Boundedness.} This property requires the measurement to have constant upper bounds $c_{\text{max}}$ and lower bounds $c_{\text{min}}$ for any node attributes. Formally, we have $c_{\text{min}} \leq  h_{\text{edge}}(\mathbf{x}_{i}, \mathbf{x}_{j}) < c_{\text{max}}, \forall{\mathbf{x}_{i}, \mathbf{x}_{j}} \in \mathcal{R}^{d} $. 
Unbounded measurements may be overly sensitive to a few attributes with extreme values. For example, overemphasizing the frequency of sending messages may misidentify highly engaged users as spammers.

    \item  \textbf{Minimal Agreement.} In GFD, if two users share identical profiles, the heterophily between them should reach its minimum value without any other information. This property is defined as minimal agreement, i.e., $c_{\text{min}} = h_{\text{edge}}(\mathbf{x}_{i}, \mathbf{x}_{j}) \iff \mathbf{x}_{i}=\mathbf{x}_{j}$.
    \item \textbf{Monotonicity.} To precisely distinguish heterophilic connections from normal edges, a promising heterophily metric should increase as the distance between attributes increases. While it is challenging to model it without prior knowledge of attribute distribution, we provide a definition that focuses on local inconsistency: 
    \begin{equation}
\begin{split}
\forall \mathbf{x}_{i} \neq \mathbf{x}_{j}\text{, }1\leq k \leq d&\text{, we have:} \\
\text{If } x_{i,k} > x_{j,k}
\text { then: }  & \frac{\partial h_{\text{edge}}(\mathbf{x}_{i}, \mathbf{x}_{j}) }{\partial x_{i,k}} > 0\\
\text{Otherwise: } & \frac{\partial  h_{\text{edge}}(\mathbf{x}_{i}, \mathbf{x}_{j}) }{\partial x_{i,k}} < 0\text{.}\\
\end{split}
\end{equation}

  \item \textbf{Equal Attribute Tolerance.} Fraudsters often mimic the attributes of benign users to evade detection. 
To avoid bias from globally shared attributes, similar to empty class tolerance~\cite{platonov2024characterizing}, we define equal attribute tolerance as:
$\forall{k}, h_{\text{edge}}(\mathbf{x}_{i}, \mathbf{x}_{j})=h_{\text{edge}}([\mathbf{x}_{i}^T||k]^T, [\mathbf{x}_{j}^T||k]^T)$, where $||$ denotes the concatenate operation. 
\end{itemize}

Based on the aforementioned properties, we analyze the existing unsupervised metrics and summarize their properties in Table~\ref{table:metrics}. Detailed proof and examples are provided in Appendix B. While these metrics can estimate heterophily to some extent, 
they fail to capture important clues essential for GFD. 
For example, cosine distance and attribute heterophily rate may overlook scale when attributes are sparse, diminishing their effectiveness in detecting accounts with numerous fake followers. Conversely, Euclidean distance may overemphasize attributes with strong disparity, such as the number of posts or likes, thereby neglecting the semantic differences in posts. Such insufficiencies limit their ability to estimate heterophily in GFD tasks, particularly when fraudsters attempt to conceal themselves among benign users.

\noindent \textbf{Harmonic Label-free Heterophily.} With regard to the limitations of existing measures, an ideal heterophily metric for GFD should be robust across diverse attribute distributions in various datasets, while remaining sensitive to suspicious attributes to effectively identify fraudulent nodes. To bridge the gap, we propose an attribute-based heterophily metric termed \textbf{HA}rmonic \textbf{L}abel-free heter\textbf{O}phily (\ourhetero), which can be written by:
\begin{equation}
\label{equation:Hedge}
\text{HALO}(\mathbf{x}_i, \mathbf{x}_j)
= \frac{\|\mathbf{\hat{x}}_i - \mathbf{\hat{x}}_j\|_2}{(\|\mathbf{\hat{x}}_i\|_2^2 + \|\mathbf{\hat{x}}_j\|_2^2+\epsilon)^\frac{1}{2}}, 
\end{equation}

\noindent where $\mathbf{\hat{x}}_i= \text{abs}(\mathbf{x}_i - \mathbf{x}_j) \odot \mathbf{x}_i$ and  $\mathbf{\hat{x}}_j= \text{abs}(\mathbf{x}_i - \mathbf{x}_j) \odot \mathbf{x}_j$. Here, $\text{abs}(\cdot)$ and $\odot$ denote element-wise absolute value and multiplication, respectively, and $\epsilon$ is a small positive number to avoid division by zero. To calculate \ourhetero, we first rescale attributes based on their pairwise absolute distance, and then compute the normalized Euclidean distance between the rescaled attributes to ensure HALO is bounded. 

With theoretical analysis, we prove that \ourhetero strictly satisfies boundedness, minimal agreement and equal attributes tolerance, and satisfies monotonicity under relaxed constraints. The detailed proofs are included in Appendix C. Satisfying boundedness ensures \ourhetero will not overemphasize a few attributes, while minimal agreement and equal attributes tolerance ensure the metric remains well-defined and robust across various GFD datasets with different attribute distributions. We relax strict monotonicity to enhance sensitivity to attributes that may camouflage fraudsters, aligning with our objective of using heterophily to guide unsupervised GFD. 

To compute node-level heterophily, we simply average the edge-level HALO between a node and its neighbors:
\begin{equation}
\label{equation:Hnode}
h_i = \text{HALO}(v_i) = \frac{\sum_{v_j \in \mathcal{N}(v_i)}{\text{HALO}(\mathbf{x}_i, \mathbf{x}_j)}}{|\mathcal{N}(v_i)|},
\end{equation}
\noindent which provides a straightforward and intuitive measure of local heterophily within the graph. While it is possible to design more theoretically grounded node-level heterophily measures~\cite{platonov2024characterizing}, we leave this as a direction for future research. 

\subsection{Alignment-based Graph Fraud Detection}
To effectively utilize \ourhetero to enhance the performance of unsupervised GFD, we propose an alignment-based graph fraud detection module, which consists of a joint MLP-GNN architecture with ranking loss and asymmetric alignment loss. 
Specifically, we first learn node embeddings from the MLP encoder to estimate local inconsistency scores as indicators of potential fraudsters.
Subsequently, we align the order of local inconsistency scores to the order of computed heterophily \ourhetero through the ranking loss, which reduces the influences of noise. 
Next, we use the GNN encoder to capture the graph structural information, and design the asymmetric alignment loss for aligning the neighbor inconsistency distribution between the MLP and the GNN encoders.
The asymmetric alignment loss not only effectively integrates structure information into the MLP encoder, but also alleviates the feature-smoothing effects of GNN.

\noindent \textbf{Local Inconsistency Estimation.} In the first step, we obtain the node embedding using an MLP encoder, which can be written as  
    $\mathbf{E} = \text{MLP}(\mathbf{X})$, where $\mathbf{E}=\{\mathbf{e}_i\}_{i=1}^{n}$ and $\mathbf{e}_i$ represents the embedding of node $v_i$. 
Utilizing an MLP as an encoder ensures that the embeddings maintain their distinctiveness, thereby preventing the discriminative indicators of fraudsters from being smoothed out by message passing. After that, we acquire the local inconsistency score $s_i$ for node $v_i$ by calculating the average cosine similarity between the node and its neighbors:
\begin{equation}
s_i = -\frac{\sum_{v_j \in \mathcal{N}(v_i)}{\text{cos}(\mathbf{e}_i, \mathbf{e}_{{j}})}}{|\mathcal{N}(v_i)|},
\end{equation}
where $\text{cos}(\cdot,\cdot)$ denotes the cosine similarity. This local inconsistency score effectively distinguishes suspicious attributes from benign ones, enabling us to effectively identify fraudsters. Therefore, we use it as the final anomaly score, i.e., $s^{\text{fraud}}_i= s_i$, ensuring effective unsupervised GFD.

\noindent \textbf{Heterophily-guided Ranking Loss.} While \ourhetero provides useful clues for GFD, directly utilizing the heterophily as the fraud score or regression target may yield sub-optimal results. As the heterophily is estimated from attributes, it often contains redundant information, such as noise or variation on attributes that are irrelevant to GFD. Such unreliable information may introduce noisy signals and reduce overall performance~\cite{burges2005learning}. 
Moreover, heterophily only captures local inconsistencies, whereas global information is also essential for detecting the most malicious entities for unsupervised GFD problem~\cite{jin2021anemone}.
To address this, we incorporate concepts from information retrieval~\cite{burges2005learning} and employ a ranking loss to align the scores predicted by the learnable GFD model with the order of estimated heterophily:
\begin{equation}
\text{\fontsize{8.5}{9}\selectfont$
\mathcal{L}_{\text{rank}}^{\text{+}} = \frac{1}{|\mathcal{B}|^2}\sum_{v_i \in \mathcal{B}}\sum_{v_j \in \mathcal{B}}\big(\mathbf{1}[h_j>h_i](s_i-s_j)+\text{log}(1+\text{e}^{-(s_i-s_j)})\big),
$}
\end{equation}

\begin{table*}[t!]
\centering
\setlength{\tabcolsep}{1mm}
\fontsize{9pt}{9pt}\selectfont
\begin{tabular}{l | c c | c c | c c | c c | c c| c c } 
 \toprule
 \multirow{2}*{{{Methods}}} & \multicolumn{2}{c|}{Amazon}& \multicolumn{2}{c|}{Facebook}& \multicolumn{2}{c|}{Reddit}& \multicolumn{2}{c|}{YelpChi}& \multicolumn{2}{c|}{AmazonFull}&\multicolumn{2}{c}{YelpChiFull}\\
 \cmidrule(r){2-13}
 & AUROC & AUPRC & AUROC & AUPRC & AUROC & AUPRC & AUROC & AUPRC & AUROC & AUPRC & AUROC & AUPRC \\
 \midrule
DOMINANT& 0.4845  & 0.0589  & 0.4372  & 0.0194  & 0.5588  & 0.0371  & 0.3993  & 0.0385  & 0.4496  & 0.0574  & \multicolumn{2}{c}{OOM}\\
CoLA & 0.4734  & 0.0683  & 0.8366  & 0.2155  & \pmb{0.6032}  & 0.0440  & 0.4336  & 0.0448  & 0.2110  & 0.0536  & \underline{0.4911}  & \underline{0.1419}  \\
ANEMONE & 0.5757  & 0.1054  & 0.8385  & 0.2320  & 0.5853  & 0.0417  & 0.4432  & 0.0473  & 0.6056  & \underline{0.2396}  & 0.4601  & 0.1305  \\
GRADATE & 0.5539  & 0.0892  & 0.8809  & 0.3560  & 0.5671  & 0.0389  & \multicolumn{2}{c|}{OOM}& \multicolumn{2}{c|}{OOM}& \multicolumn{2}{c}{OOM}\\
ADA-GAD & 0.5240  & 0.1108  & 0.0756  & 0.0122  & 0.5610  & 0.0382  & \multicolumn{2}{c|}{OOM}& \multicolumn{2}{c|}{OOM}& \multicolumn{2}{c}{OOM}\\
GADAM& 0.6167  & 0.0857  & \underline{0.9539}  & \underline{0.3630}  & 0.5809  & \underline{0.0465}  & 0.4177  & 0.0423  & 0.4457  & 0.0566  & 0.4797  & 0.1351  \\
TAM  & \underline{0.7126}  & \underline{0.2915}  & 0.8895  & 0.1937  & 0.5748  & 0.0438  & \underline{0.5473}  & \pmb{0.0780}  & \underline{0.6442}  & 0.2188  & \multicolumn{2}{c}{OOM}\\

 \midrule
\textbf{\ourmethod (ours)}& \pmb{0.8516} & \pmb{0.6672} & \pmb{0.9760} & \pmb{0.3674} &  \underline{0.5906} & \pmb{0.0511} & \pmb{0.6013} &  \underline{0.0708} & \pmb{0.8892} & \pmb{0.7668} & \pmb{0.5767} & \pmb{0.1869}\\

 \bottomrule
\end{tabular}
\caption{Unsupervised GFD performance comparison between our proposed \ourmethod and baselines with AUROC and AUPRC. The best and second-best results are in \textbf{bold} and \underline{underlined}, respectively. OOM indicates out-of-memory on a 24GB GPU.}

\label{table:results}
\end{table*}
\noindent where $\mathcal{B}$ represents a batch of sampled nodes, and $\mathbf{1}[\cdot]$ is the indicator function which returns 1 when the condition is satisfied and 0 otherwise. By avoiding direct mapping the scores to specific values, the ranking loss ensures the robustness of the GFD model and mitigates the influence of irrelevant information and bias~\cite{burges2005learning}. Furthermore, the pairwise ordering relation allows comparisons between non-adjacent nodes, providing valuable global insights and enhancing the detection of fraudulent entities.

In addition to the order of heterophily, we highlight another implicit but important prior inspired by the homophily assumption~\cite{zheng2022graph}: the local inconsistency score ${s}_i$ between a node and its neighbor should be smaller than the inconsistency score between a node and its non-neighbor ${s}_{i}^{\text{-}}$. 
Therefore, by integrating this hidden constraint into the ranking loss, we can obtain:
\begin{equation}
\mathcal{L}_{\text{rank}}^{\text{-}} = \frac{1}{|\mathcal{B}|}\sum_{v_i \in \mathcal{B}}\big(s_i-{s}^{\text{-}}_i+\text{log}(1+\text{e}^{-(s_i-{s}^{\text{-}}_i)})\big),
\end{equation}
\begin{equation}
{s}^{\text{-}}_i = -\frac{\sum_{v_j \in \mathcal{B} \backslash \mathcal{N}(v_i)}{\text{cos}(\mathbf{e}_i, \mathbf{e}_{j})}}{|\mathcal{B} \backslash \mathcal{N}(v_i)|}, 
\end{equation}

\noindent where $\text{cos}(\cdot,\cdot)$ denotes the cosine similarity. Finally, we use the average of $\mathcal{L}_{\text{rank}}^{\text{+}}$ and $\mathcal{L}_{\text{rank}}^{\text{-}}$ as the final rank loss: 
\begin{equation}
\label{rankingloss}
\mathcal{L}_{\text{rank}} = \frac{\mathcal{L}_{\text{rank}}^{\text{+}} + \mathcal{L}_{\text{rank}}^{\text{-}}}{2}.
\end{equation}

\noindent \textbf{Asymmetric  Alignment Loss.} Apart from attributes, the graph structure offers valuable information for unsupervised GFD. To leverage this without obscuring fraud clues through message passing, we introduce an auxiliary GNN after the MLP encoder. 
We then employ an asymmetric alignment loss to extract structural knowledge from GNN to MLP by aligning their neighbor inconsistency distribution (NID). 

In detail, we first propagate and project the output embedding $\mathbf{E}$ from the MLP encoder using a GNN layer:
\begin{equation}
   \bar{\mathbf{E}} = \text{GNN}(\mathbf{E}, \mathbf{A}),
\end{equation}

\noindent where $\mathbf{A}$ is the adjacent matrix. 
After that, we align the NID (i.e. the distribution of rescaled cosine similarity between the embedding of $v_i$ and its neighbors) of $\mathbf{E}$ and $\bar{\mathbf{E}}$ using an asymmetric alignment loss:
\begin{equation}
\text{\fontsize{8.5}{9}\selectfont$
\begin{aligned}
\mathcal{L}_\text{align} & = \frac{1}{|\mathcal{V}|}\sum_{v_i \in \mathcal{V}}\text{KLD}\big(\mathbf{s}^{\text{NID}}_{i}| \text{sg}[\bar{\mathbf{s}}^{\text{NID}}_{i}]\big)\\
& = \frac{1}{|\mathcal{V}|}\sum_{v_i \in \mathcal{V}} \frac{1}{|\mathcal{N}(v_i)|} \sum_{v_j \in \mathcal{N}(v_i)} 
\text{sg}[\bar{s}_{i,j}]\big(\text{log}(\text{sg}[\bar{s}_{i,j}])-\text{log}(s_{i,j})\big),
\end{aligned}$}
\end{equation}
\noindent where $\text{sg}[\cdot]$ denotes stop gradient operation, $\text{KLD}(\cdot|\cdot)$ denotes the Kullback-Leibler Divergence, and $s_{i,j}$, $\bar{s}_{i,j}$ represent the rescaled cosine similarity (rescaled to $[0, 1]$) between the embeddings of node $i$ and $j$ for MLP and GNN, respectively. The  $\mathcal{L}_\text{align}$ transfers the structural information captured by the GNN to the MLP by minimizing the KLD. 
This approach not only enhances the ability of the model to detect fraudulent entities, but also reduces the feature-smooth effect introduced by solely using GNN. Note that we employ an asymmetry loss with a stop gradient to prevent the GNN from degrading into an identical mapping. 

Similar to the training of the MLP encoder, we also compute the ranking loss $\bar{\mathcal{L}}_\text{rank}$ for the GNN branch following Eq.~(\ref{rankingloss}), which results in our overall objective with a hyper-parameter $\alpha$ to control the strength of alignment:
\begin{equation}
   \mathcal{L} = \mathcal{L}_\text{rank} +  \bar{\mathcal{L}}_\text{rank} + \alpha \mathcal{L}_\text{align}. 
\end{equation}

\noindent \textbf{Complexity Analysis.} To sum up, the heterophily estimation, training, and testing complexities of \ourmethod are $\mathcal{O}(m d)$, $\mathcal{O} \big((n d + m d_e + n|\mathcal{B}|)d_e\big )$~(per epoch), and $\mathcal{O}\big( (n d + m) d_e \big)$, respectively, where $d_{e}$ is the embedding dimension. Detailed complexity analysis can be found in Appendix D.

\section{Experiments}
\subsection{Experimental Settings}
\noindent \textbf{Datasets.} 
We conduct experiments on six public real-world GFD datasets, covering domains ranging from social networks to e-commerce, including Amazon~\cite{dou2020enhancing}, Facebook~\cite{xu2022contrastive}, Reddit, YelpChi~\cite{kumar2019predicting}, AmazonFull~\cite{mcauley2013amateurs}, and YelpChiFull~\cite{rayana2015collective}. 
Following~\cite{he2024ada, qiao2024truncated}, we convert all graphs to homogeneous undirected graphs in our experiments.

\begin{figure*}
\centering
\subfigure[Parameter sensitivity of $\alpha$.]{
\label{fig:sensitivity_alpha:a}
\includegraphics[width=.27\linewidth]{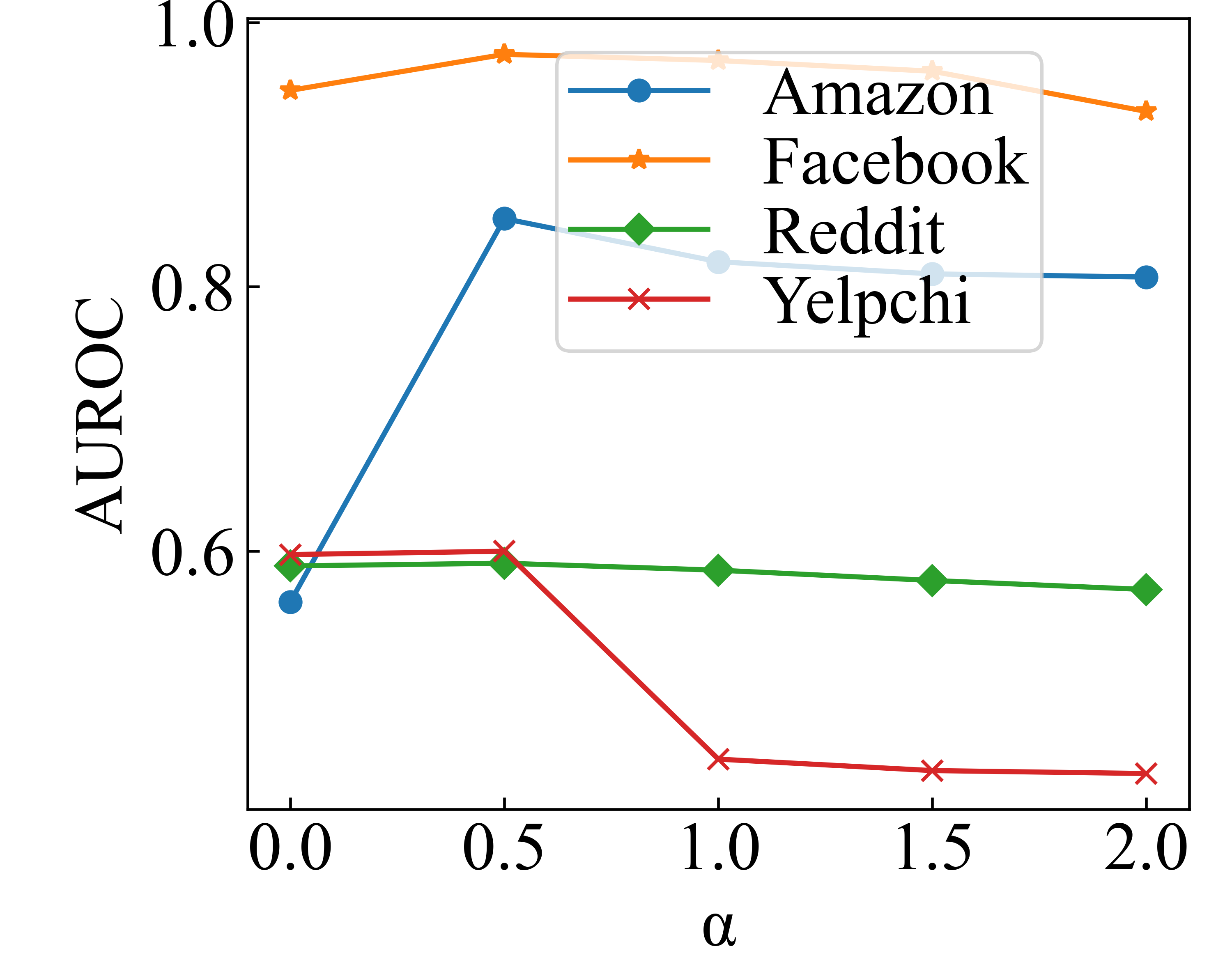}
}
\hfill
\subfigure[Visualization of fraud scores.]{
\label{fig:sensitivity_alpha:b}
\includegraphics[width=.27\linewidth]{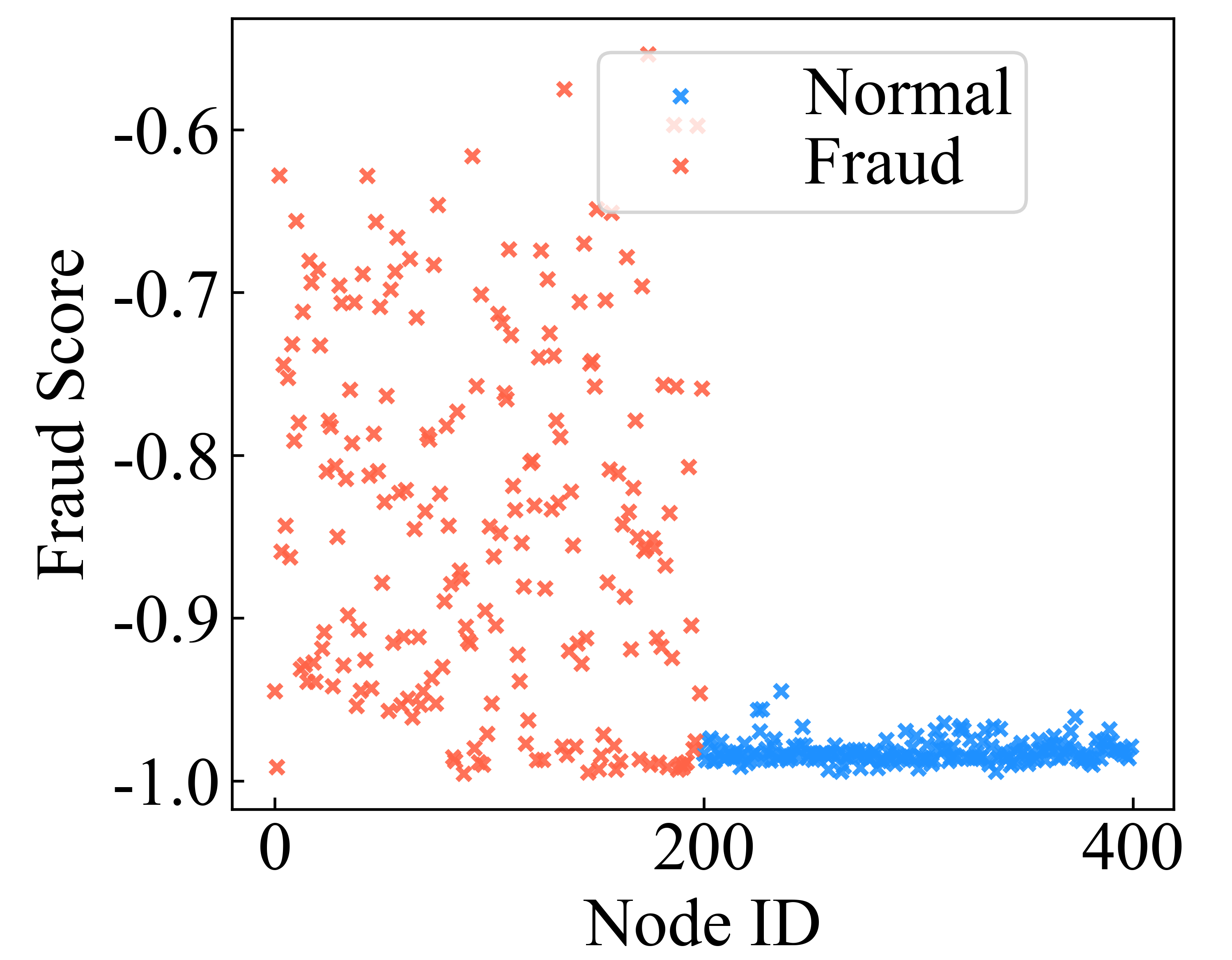}
}
\hfill
\subfigure[Efficiency in terms of training time.]{
\label{fig:sensitivity_alpha:c}
\includegraphics[width=.27\linewidth]{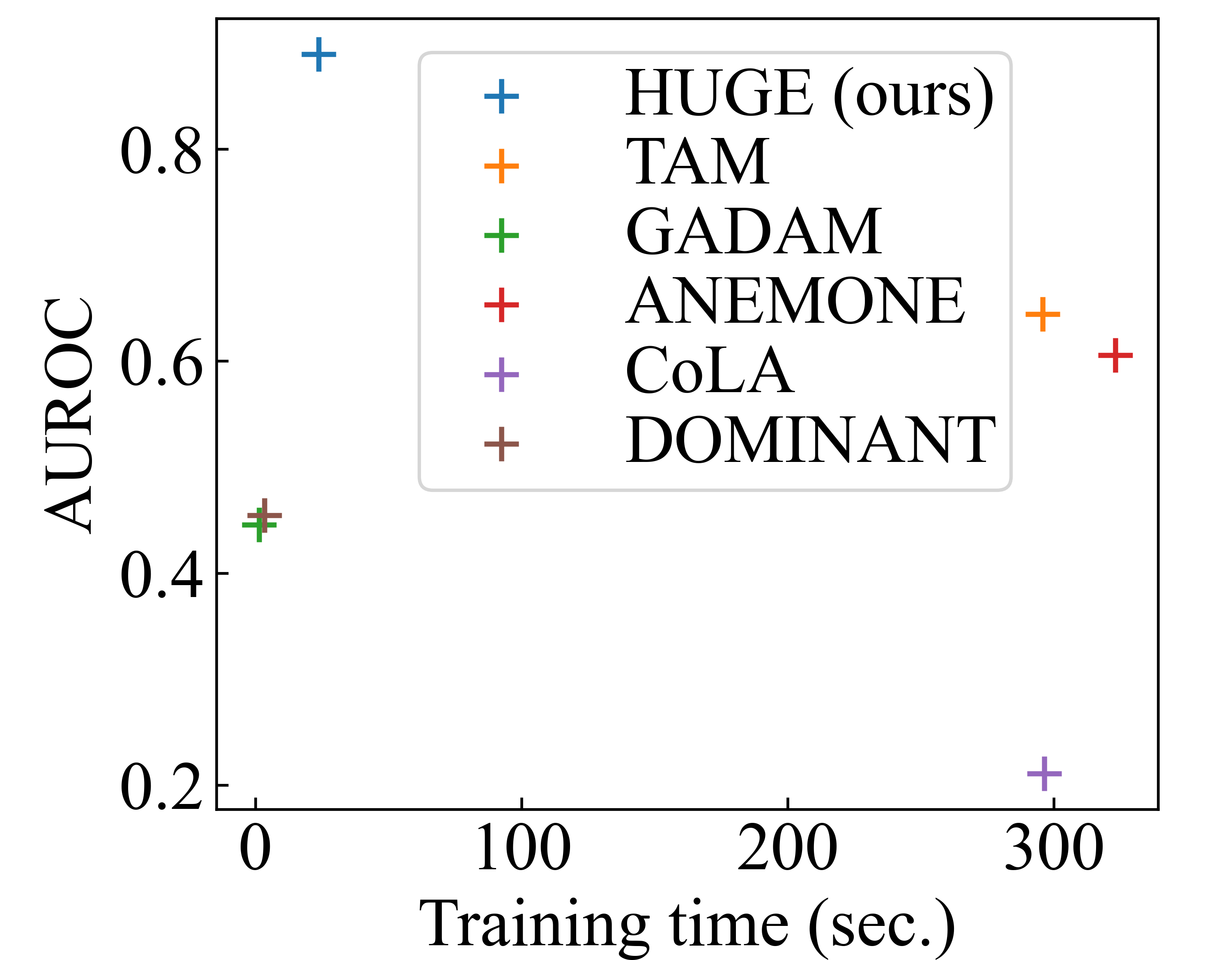}
}
\caption{Visualization of results. (a) Parameter sensitivity w.r.t. $\alpha$ of the proposed \ourmethod. (b) Fraud score visualization on AmazonFull dataset. (c) AUROC and training time of \ourmethod and baselines on AmazonFull dataset.}
\label{fig:sensitivity_alpha}
\end{figure*}

\noindent \textbf{Baselines and Evaluation Metrics.} 
We compare our method with 7 state-of-the-art (SOTA) unsupervised GAD methods: DOMINANT~\cite{ding2019deep}, CoLA~\cite{liu2021anomaly}, ANEMONE~\cite{jin2021anemone}, GRADATE~\cite{duan2023graph}, ADA-GAD~\cite{he2024ada}, GADAM~\cite{chen2024boosting} and TAM~\cite{qiao2024truncated}. The area under the receiver operating characteristic curve (AUROC) and the area under the precision-recall curve (AUPRC) are used as the evaluation metrics. We report the average performance over five runs with different random seeds. 
The implementation details, hyper-parameters used in our experiments, and more results are presented in Appendix E and F.

\subsection{Experimental Results}

\noindent \textbf{Performance Comparison.} The comparison results on six real-world GFD datasets are reported in Table \ref{table:results}. 
From the results, we make the following key observations: 
\ding{182}~\ourmethod outperforms SOTA methods in most datasets in terms of both metrics, except for AUROC on Reddit and AUPRC on YelpChi. These results demonstrate the effectiveness of \ourmethod in diverse real-world GFD scenarios. Moreover, while existing methods show unstable performance across different datasets, \ourmethod consistently achieves competitive results. 
\ding{183}~Among the baselines, the methods that consider heterophily (ADA-GAD, GADAM, TAM) usually outperform other baselines, demonstrating the benefits of mitigating heterophily for unsupervised GFD. However, their performances are constrained by their heterophily estimations. In contrast, our proposed heterophily measurement HALO shows expressive performance. 
\ding{184}~The scalability of several baselines is challenged by large-scale real-world GFD datasets with out-of-memory (OOM) issues, e.g., YelpChiFull. In contrast, \ourmethod exhibits better scalability, highlighting its potential in handling real-world large-scale data from online services. 

\noindent \textbf{Ablation Study.} To examine the contribution of key design in \ourmethod, we conduct ablation studies on heterophily measures and critical model components. 

To evaluate the effectiveness of our label-free heterophily \ourhetero (w/ HALO), we replace it with other heterophily measures, i.e. Euclidean distance (w/ Euc. Dist.), cosine distance (w/ Cos. Dist.), attribute heterophily rate (w/ AHR), as illustrated in the upper part of Table \ref{table:ablation2}. We can observe that our \ourhetero achieves superior performances compared to other heterophily measures, which verifies its effectiveness for unsupervised GFD. 
In addition, while some other measures demonstrate promising results on certain datasets, none of them consistently excels across all datasets, highlighting their limitations in guiding GFD models.

\begin{table}[t!]
\setlength{\tabcolsep}{1mm}
\fontsize{9pt}{9pt}\selectfont
\centering
\begin{tabular}{l | c c c c c c } 
 \toprule
 Measures & Amazon & Facebook & AmazonFull & YelpChiFull  \\
 \midrule
  \begin{tabular}{@{}c@{}}\textbf{\ourmethod} \\ \textbf{(w/ HALO)}\end{tabular}   & \textbf{0.8516} & \textbf{0.9760} & \textbf{0.8892} & \textbf{0.5767} \\

 \midrule
w/ Euc. Dist.\ & 0.7128 & 0.9251 & 0.8746 & 0.5763 \\
w/ Cos. Dist. & 0.7018 & 0.9668 & 0.8735 & 0.5762  \\
w/ AHR & 0.8171 & 0.9438 & 0.8622 &  0.5765\\
\midrule
w/o Alignment\ & 0.5614 & 0.9487 & 0.8746  & 0.5295 \\
w/o GNN & 0.5548 & 0.9401 & 0.7005  & 0.5533  \\
 \bottomrule
\end{tabular}
\caption{Ablation study of key designs in \ourmethod. 
}

\label{table:ablation2}
\end{table}

We compare \ourmethod with two variants that exclude the corresponding components, i.e., the asymmetric alignment loss (w/o Alignment) and GNN branch (w/o GNN), to verify the effectiveness of these key designs.
The results in the lower part of Table~\ref{table:ablation2} demonstrate that \ourmethod, with all components, achieves the best performance among all datasets, illustrating the effectiveness and robustness of each module.  Meanwhile, the asymmetric alignment loss provides the major contribution, when removing it results in a significant performance drop. Moreover, while solely attaching the GNN model may lead to a performance drop (as illustrated in YelpChiFull) due to feature-smoothing effects, our alignment loss can overcome the issue by extracting the structural information to the MLP encoder.

\noindent \textbf{Sensitivity Analysis.} In order to evaluate the sensitivity of \ourmethod to the hyper-parameter $\alpha$, we adjust its value across $\{0.0, 0.5, 1.0, 1.5, 2.0\}$. The results are illustrated in Figure~\ref{fig:sensitivity_alpha:a}. Overall, \ourmethod is not sensitive to variation in $\alpha$ on Facebook and Reddit, and slightly sensitive on Amazon and YelpChi. Moreover, \ourmethod achieves the best performance at $\alpha=0.5$, with a noticeable drop when $\alpha=0$ on Amazon and Facebook. These findings suggest \ourmethod performs best when neighbor information and ego information are balanced. Additionally, the small performance gap between $\alpha=0$ and $\alpha=1$ on Reddit and YelpChi shows that these datasets require less neighbor information for effective fraud detection.

\noindent \textbf{Visualization.} The distribution of learned fraud scores is visualized in Figure~\ref{fig:sensitivity_alpha:b} by plotting a subset of scores. We observe that the fraudsters tend to have higher fraud scores, while benign nodes receive lower scores. The gap between the distributions of fraud and benign nodes highlights the effectiveness of \ourmethod. Moreover, the score distribution of fraudsters is more diverse than the score distribution of normal nodes, indicating the diversity of frauds in real-world unsupervised GFD scenarios. Nevertheless, \ourmethod still achieves favorable results. 

\noindent \textbf{Efficiency Comparison.} To compare the efficiency of our method against baselines, we plot their training time against AUROC. As shown in Figure~\ref{fig:sensitivity_alpha:c}, \ourmethod obtains the best performance while maintaining superior efficiency. Although DOMINANT and GADAM can be trained faster, they fail to effectively detect fraudsters, performing worse than random guesses (AUROC=0.5). The high running efficiency indicates the potential of applying \ourmethod for real-time detection in practical scenarios. 

\section{Conclusion}
In this paper, we propose a novel unsupervised graph fraud detection (GFD) method termed \ourmethod, which effectively alleviates the negative effects of heterophily introduced by fraudsters. To address the challenge of estimating heterophily without labeled annotations, we design an effective label-free heterophily metric, \ourhetero. Then, we use \ourhetero to guide the training of our proposed model through a ranking loss. Additionally, we introduce an asymmetric alignment loss to extract useful structural information while mitigating the smoothing effect introduced by GNNs. Comprehensive experiments showcase the superior performance of \ourmethod over baseline methods across diverse datasets. 

\section{Acknowledgments}
This research was partly funded by Australian Research Council (ARC) under grants FT210100097 and DP240101547 and the CSIRO – National Science Foundation (US) AI Research Collaboration Program.

\bibliography{aaai25}

\appendix
\section{Appendix A: Related Work in Detail}
\label{Appendix:A}
\subsection{Graph Fraud Detection}
Graph fraud detection (GFD) aims to identify fraudulent activities from graph-structured real-world systems, including financial fraud~\cite{motie2023financial}, spamming~\cite{deng2022markov}, and fake reviews~\cite{yu2022graph}. Various studies have proposed different models to resolve the problem. For example, GraphConsis~\cite{liu2020alleviating} tackles the inconsistency problem by computing consistency between nodes, while CARE-GNN~\cite{dou2020enhancing} designs a similarity-aware neighbor selector to address camouflage behaviors. Similarly, PC-GNN~\cite{liu2021pick} proposes a neighbor sampler to balance label distribution. Although these GNN-based methods have achieved promising results, they suffer from heterophily caused by the camouflages problem~\cite{dou2020enhancing}, i.e., benign nodes tend to have heterophilic connections with normal nodes to make them indistinguishable from the majority.

Recent studies try to develop advanced GFD approaches with the consideration of heterophily. With the help of labels, a variety of techniques have been employed to mitigate the impact of heterophily. H\textsuperscript{2}-FDetector~\cite{shi2022h2} trains a classifier to identify connection types, guiding attention-based aggregation, whereas GHRN~\cite{gao2023addressing} prunes heterophily edges instead. BWGNN~\cite{tang2022rethinking} examines the right shift phenomenon to design an expressive encoder. GAGA~\cite{wang2023label} combines group aggregation and learnable encoding to fully utilize label information. GDN~\cite{gao2023alleviating} reveals the heterophily shift across training and testing datasets and mitigates the issue through feature separation, while ConsisGAD~\cite{chen2024consistency} directly utilizes homophily patterns for distinguishing normal and anomalous nodes. SEC-GFD~\cite{xu2024revisiting} employs spectrum analysis to aggregate frequent bands separately, and PMP~\cite{zhuo2024partitioning} resolved the problem in spatial domains by aggregating homophilic and heterophilic neighbors independently. 

Although these methods achieve promising results in mitigating heterophily, their reliance on labels limits their applicability in scenarios where such labels are unavailable. Due to the significant costs of annotating labels in GFD-related scenarios~\cite{sehwag2021ssd}, there is a compelling need to create unsupervised GFD methods.

\subsection{Graph Anomaly Detection}
Graph anomaly detection (GAD) is a broader concept than GFD, aiming to identify not only fraudsters but also any rare and unusual patterns that significantly deviate from the distribution of the majority in graph data. Therefore, GAD techniques can be directly applied to GFD, particularly within unsupervised learning settings. Due to the broad scope of GAD and the difficulty in obtaining real-world anomalies, many unsupervised GAD methods have been designed and evaluated on several datasets with injected anomalies. For example, DOMINANT~\cite{ding2019deep} uses reconstruction error to estimate the abnormality of nodes, while CoLA~\cite{liu2021anomaly} employs contrastive learning to compute ego-neighbor disagreement as anomaly score, which is adapted by later works~\cite{jin2021anemone, zheng2021generative, duan2023graph, duan2023arise, pan2023prem}.

In spite of the decent performances they have achieved, these methods rely on the homophily assumption, which limits their applications since heterophily is a ubiquitous property in graphs~\cite{zheng2022graph}. Recent studies explore this insufficiency and suggest using estimated anomaly scores to mitigate the negative impacts of heterophily, like dropping edges~\cite{he2024ada, qiao2024truncated} or adjusted message passing~\cite{chen2024boosting}. 
However, their understanding of heterophily is fundamental and lacks a systematic methodology for defining the metric. Moreover, their estimation of heterophily heavily relies on embeddings computed by message-passing GNNs, which may be unreliable in the first place~\cite{zhu2022does}. 
Consequently, their estimated anomaly scores for GFD datasets could be poor, even worse than randomly generated scores~\cite{GLOD}. To fill the gap, our proposed \ourmethod first introduces a simple yet effective heterophily measure with a set of desired properties to address the aforementioned insufficiency, which can be further used to guide the learning of the downstream unsupervised GFD model.

\appendix
\section{Appendix B: Proof of Existing Unsupervised Edge-level Heterophily Metrics}
In this section, we validate the statements listed in Table~1 by offering mathematical proofs or presenting counter-examples to illustrate their limitations in Graph Fraud Detection (GFD).

\subsection{Euclidean Distance}

\begin{theorem}
Euclidean distance is unbounded. It satisfies minimal agreement, monotonicity and equal attribute tolerance. 
\end{theorem}
\begin{proof}
It is trivial to prove that the Euclidean distance is unbounded and satisfies minimal agreement, as the Euclidean distance between two identical vectors is zero and a zero distance implies the vectors are identical. 

To prove equal attribute tolerance, we have: 
\begin{equation}
\label{equation:l2:1}
\tag{A.1}
\begin{split}
     \|[\mathbf{x}_{a}^T\|k]^T - [\mathbf{x}_{b}^T\|k]^T\|_2 
    = & \sqrt{\sum_{i=0}^{d} (x_{a,i}-x_{b,i})^2 + (k-k)^2}\\
    = & \sqrt{\sum_{i=0}^{d} (x_{a,i}-x_{b,i})^2}\\
    = & \|\mathbf{x}_{a}-\mathbf{x}_{b}\|_2. 
\end{split}
\end{equation}

Hence, the Euclidean distance satisfies equal attribute tolerance.

To prove monotonicity, we first compute the derivative of the Euclidean distance with respect to (w.r.t.) \( \mathbf{x}_{a, i} \):
\begin{equation}
\tag{A.2}
\begin{split}
f^{'} = & \frac{\partial (\|\mathbf{x}_{a}-\mathbf{x}_{b}\|_2)}{\partial x_{a, i}}\\
= & \frac{\partial \big(\sqrt{\sum_{j=1}^d (x_{a, j}-x_{b, j})^2}\big )}{\partial x_{a, i}}\\
= & \frac{x_{a, i}-x_{b, i}}{\|\mathbf{x}_{a}-\mathbf{x}_{b}\|_2}.
\end{split}
\end{equation}
Hence, the derivative $f^{'}$ is positive when $x_{a, i}-x_{b, i} > 0$, and $f^{'}$ is negative otherwise. Therefore, Euclidean distance satisfies the conditions for monotonicity.
\end{proof}

\subsection{Cosine Distance}

\begin{theorem}
Cosine distance is bounded. It does not satisfies minimal agreement, monotonicity and equal attribute tolerance. 
\end{theorem}

\begin{proof}
The range of cosine distance is $[-1, 1]$, which proves its boundedness. Meanwhile, the cosine distance between parallel vectors with different norm is always $-1$, indicating that it does not satisfy minimal agreement or  monotonicity. Furthermore, it fails to meet the criteria of equal attribute tolerance. For example, $\text{cd}([1, 2]^T, [3, 4]^T) = 0.016$, while $\text{cd}([1, 2, 5]^T, [3, 4, 5]^T) = 0.070$, where $\text{cd}(\cdot,\cdot)$ denotes cosine distance. 
\end{proof}

\subsection{Attribute Heterophily Rate}

\begin{theorem}
The attribute heterophily rate satisfies both boundedness and minimal agreement. It does not satisfies monotonicity and equal attribute tolerance. 
\end{theorem}
\begin{proof}
Given that the range of the indicator function $\mathbf{1}$ is $\{0, 1\}$, it is straightforward to prove that the attribute heterophily rate is bounded by [0, 1] and satisfies minimal agreement. However, as the indicator function only checks whether the two values are identical, it fails to satisfy monotonicity. For equal attribute tolerance, we have: 
\begin{equation}
\tag{A.3}
\begin{split}
 \text{AHR}
 \begin{pmatrix}
    \begin{bmatrix}
         \mathbf{x}_{a} \\
         k \\
    \end{bmatrix}
    , 
    \begin{bmatrix}
         \mathbf{x}_{b} \\
         k \\
    \end{bmatrix}
 \end{pmatrix} 
= &\frac{\sum^{d}_{i=1} { \mathbf{1}[x_{a, i}\neq x_{b, i}]} + \mathbf{1}[k\neq k]}{d + 1}\\
=&\frac{d\text{AHR}(\mathbf{x}_{a}, \mathbf{x}_{b})+1}{d + 1},
\end{split}
\end{equation}
where AHR$(\cdot, \cdot)$ denotes attribute heterophily rate. Hence, the attribute heterophily rate does not satisfy equal attribute tolerance unless $\text{AHR}(\mathbf{x}_{a}, \mathbf{x}_{b}) = 1$, which occurs only when $\mathbf{x}_{a}=\mathbf{x}_{b}$. 
\end{proof}

\appendix
\section{Appendix C: Proof of Harmonic Label-Free Heterophily}
In this section, we provide a detailed proof to show that our edge-level harmonic label-free heterophily (\ourhetero) satisfies boundedness, minimal agreement and equal attribute tolerance, while it satisfies monotonicity under a relaxed constraint, where the length of feature vectors are constant. 

As a recap, our \ourhetero is defined as: 
\begin{equation}
\tag{A.4}
\label{equation:Hedge2}
\text{HALO}(\mathbf{x}_a, \mathbf{x}_b) = \frac{\|\mathbf{\hat{x}}_a - \mathbf{\hat{x}}_b\|_2}{(\|\mathbf{\hat{x}}_a\|_2^2 + \|\mathbf{\hat{x}}_b\|_2^2+\epsilon)^\frac{1}{2}}, 
\end{equation}
where  $\mathbf{\hat{x}}_a= \text{abs}(\mathbf{x}_a - \mathbf{x}_b) \odot \mathbf{x}_a$, $\mathbf{\hat{x}}_b= \text{abs}(\mathbf{x}_a - \mathbf{x}_b) \odot \mathbf{x}_b$ are preprocessed attributes to emphasize heterophilic patterns. The abs, $\odot$, $\epsilon$ denote element-wise absolute value, multiplication, a small positive number to keep the denominator larger than zero, respectively.

\begin{theorem}[Boundedness]
\ourhetero satisfies boundedness. 
\end{theorem}

\begin{proof}
It is trivial to prove \ourhetero has a minimum value of 0. For the proof of the upper bound, with the triangle inequality of vectors, we have:
\begin{equation}
\tag{A.5}
\begin{split}
\|\mathbf{\hat{x}}_a - \mathbf{\hat{x}}_b\|_2^2 \leq & ( \|\mathbf{\hat{x}}_a\|_2 + \|\mathbf{\hat{x}}_b\|_2 )^2\\
= & \|\mathbf{\hat{x}}_a\|_2^2 + \|\mathbf{\hat{x}}_b\|_2^2 + 2\|\mathbf{\hat{x}}_a\|_2\|\mathbf{\hat{x}}_b\|_2\\
\leq & 3(\|\mathbf{\hat{x}}_a\|_2^2 + \|\mathbf{\hat{x}}_b\|_2^2).
\end{split}
\end{equation}

Therefore, \ourhetero is bounded by 0 and $\sqrt{3}$. 
\end{proof}

\begin{theorem}[Minimal Agreement]
\ourhetero satisfies minimal agreement.
\end{theorem}
\begin{proof}
    
Since $0$ is the lower bound of \ourhetero and the denominator is always larger than zero, we only need to prove that the numerator reaches 0 if and only if the two attributes are identical. Given that the Euclidean distance satisfies minimal agreement, our \ourhetero also satisfies minimal agreement. 

\end{proof}

\begin{theorem}[Equal Attribute Tolerance]
\ourhetero satisfies equal attribute tolerance.
\end{theorem}
\begin{proof}
Since  $\mathbf{\hat{x}}_a= \text{abs}(\mathbf{x}_a - \mathbf{x}_b) \odot \mathbf{x}_a$, $\mathbf{\hat{x}}_b= \text{abs}(\mathbf{x}_a - \mathbf{x}_b) \odot \mathbf{x}_b$, we have:

\begin{equation}
\tag{A.6}
\begin{split}
\|\hat{[\mathbf{x}_{a}^T|k]^T}\|_2
=& \sqrt{\sum_{i=1}^d \hat{x}_{a,i}^2 + (k-k)k}\\
=&\|\hat{\mathbf{x}}_{a}\|_2
\end{split}    
\end{equation}
Similarly, $\|\hat{[\mathbf{x}_{b}^T|k]^T}\|_2=\|\hat{\mathbf{x}}_{b}\|_2$. By equation (\ref{equation:l2:1}), we have $\|\hat{[\mathbf{x}_{a}^T|k]^T}-\hat{[\mathbf{x}_{b}^T|k]^T}\|_2 =\|\hat{\mathbf{x}}_{a} - \hat{\mathbf{x}}_{b}\|_2$. Hence, by substituting these equations into equation (\ref{equation:Hedge2}) we prove that $\text{HALO}([\mathbf{x}_{a}^T|k]^T, [\mathbf{x}_{b}^T|k]^T) = \text{HALO}(\mathbf{x}_a, \mathbf{x}_b)$, i.e. our \ourhetero satisfies equal attribute tolerance.  

\end{proof}

\begin{theorem}[Monotonicity]
\ourhetero satisfies monotonicity where the length of feature vectors are constant. 
\end{theorem}
\begin{proof}
To facilitate the analysis of the monotonicity of \ourhetero, we have:
\begin{equation}
\tag{A.7}
\begin{split}
h &= \frac{\|\mathbf{\hat{x}}_a - \mathbf{\hat{x}}_b\|_2}{(\|\mathbf{\hat{x}}_a\|_2^2 + \|\mathbf{\hat{x}}_b\|_2^2+\epsilon)^\frac{1}{2}} \\
&= \frac{\sqrt{\sum_{k=1}^n (\hat{x}_{a,k}-\hat{x}_{b,k})^2}}
{(\|\mathbf{\hat{x}}_a\|_2^2 + \|\mathbf{\hat{x}}_b\|_2^2+\epsilon)^\frac{1}{2}}\\
&= \frac{\sqrt{(\hat{x}_{a,i}-\hat{x}_{b,i})^2 + \sum_{k=1, k \neq i}^n (\hat{x}_{a,k}-\hat{x}_{b,k})^2}}
{(\|\mathbf{\hat{x}}_a\|_2^2 + \|\mathbf{\hat{x}}_b\|_2^2+\epsilon)^\frac{1}{2}}
\end{split}
\end{equation}

To simplify the prove and analysis, we assume the length of attributes, i.e., $\|\mathbf{\hat{x}}_a\|_2$ and $\|\mathbf{\hat{x}}_b\|_2$ are constant values to relax the constraint of monotonicity, which can be achieved by feature normalization in practice. Hence, we can compute the derivative of $h$:
\begin{equation}
\tag{A.8}
\begin{split}
\frac{\partial h}{\partial \hat{x}_{a,i}} &= \frac{\hat{x}_{a, i}-\hat{x}_{b,i}}{(\|\mathbf{\hat{x}}_a\|_2^2 + \|\mathbf{\hat{x}}_b\|_2^2+\epsilon)^\frac{1}{2}\sqrt{\sum_{k=1}^n (\hat{x}_{a,k}-\hat{x}_{b,k})^2}} \\
&= m(\hat{x}_{a, i}-\hat{x}_{b,i}),
\end{split}
\end{equation}
where $m$ is a positive number. Therefore, \ourhetero satisfies monotonicity under the relaxed constraint. 

Smart fraudsters often hide themselves among benign users by manipulating their attributes to appear less distinguishable. Therefore, we rescale the attributes to uncover the hidden suspicious patterns. While this operation relaxes the constraint of monotonicity, it is better suited for unsupervised graph fraud detection scenarios. 
\end{proof}

\appendix
\section{Appendix D: Algorithm Complexity}
In this section, we discuss the time complexity of each component in \ourmethod respectively. We denote the dimension of embedding as $d_e$. For computing \ourhetero, the complexity is $\mathcal{O}(md)$. 
For the alignment-based fraud detection module, the forward pass through MLP layers and GNN layers have a complexity of $\mathcal{O}( n d d_e)$ and $\mathcal{O}( m d_e^2)$ respectively. The time complexity of computing fraud score is $\mathcal{O}(m d_e)$.
To compute the ranking loss, the complexity is $\mathcal{O}(n |B| d_e)$, and the complexity for computing asymmetric alignment loss is the same as computing fraud score. To sum up, the heterophily estimation, training  and testing complexity of \ourmethod are  $\mathcal{O}(m d)$, $\mathcal{O}\big ((n d + m d_e+n|B|)d_e\big )$ per epoch,  and $\mathcal{O}\big( (n d + m)d_e \big)$, respectively. The training algorithm of \ourmethod is summarized in Algorithm~\ref{alg:algorithm}.

\begin{table}
\renewcommand\thetable{A.1} 
\setlength{\tabcolsep}{1mm}
\fontsize{9pt}{12pt}\selectfont
\centering
\begin{tabular}{l | c   c } 
 \toprule
Params.                 & Dataset                                                                      & Range                     \\  
\midrule

\multirow{2}{*}{Epoch}   & \begin{tabular}[c]{@{}c@{}}Amazon, Facebook, \\ Reddit, YelpChi\end{tabular} & $300$ (Fixed)             \\ \cline{2-3} 
                          & AmazonFull, YelpChiFull                                                      & $\{5, 10\}$                 \\

                          \midrule

\multirow{2}{*}{lr} & \begin{tabular}[c]{@{}c@{}}Amazon, Facebook, \\ Reddit, YelpChi, \\AmazonFull\end{tabular} & $\{0.001, 0.0005, 0.0001\}$      \\ 
\cline{2-3} 
                          & YelpChiFull                                                      & $\{0.001, 0.0001, 0.00001\}$             \\  

\midrule

\multirow{2}{*}{$\alpha$} & \begin{tabular}[c]{@{}c@{}}Amazon, Facebook, \\ Reddit, YelpChi\end{tabular} & $\{0, 0.5, 1, 1.5, 2\}$     \\ 
\cline{2-3} 
                          & AmazonFull, YelpChiFull                                                      & $\{0, 1, 2, 3\}$            \\  
\bottomrule
\end{tabular}
\caption{Range of grid search.}
\label{table:parameters}
\end{table} 

\begin{table}

\renewcommand\thetable{A.2} 
\centering
\begin{tabular}{l | c c c } 
 \toprule
 Datasets & Epoch & lr & $\alpha$  \\
 \midrule
Amazon & 300 &  0.0005 & 0.5 \\
Facebook & 300 &  0.0005 & 0.5 \\
Reddit & 300 &  0.0005 & 0.5 \\
YelpChi & 300 &  0.0005 & 0.5 \\
AmazonFull & 10 &  0.0005 & 1 \\
YelpChiFull & 5 &  0.00001 & 3 \\
 \bottomrule
\end{tabular}
\caption{Hyper-parameters of \ourmethod.}
\label{table:gridsearch}
\end{table}

\begin{algorithm}[tb]
\caption{\ourmethod}
\label{alg:algorithm}
\textbf{Input}: Attributed Graph, $G=(\mathcal{V}, \mathcal{E}, \mathbf{X})$, $E$: Training epochs, $B$: Batch size, $\alpha$: Alignment parameter, lr: Learning rate.\\
\textbf{Output}: Fraud scores of all nodes $\mathbf{s}$. \\
\begin{algorithmic}[1] 
\STATE Compute the label-free heterophily $\mathbf{h}=\{h_i\}$ for each node $v_i\in \mathcal{V}$ using \ourhetero. 
\STATE Randomly initialize the parameters of the MLP and GNN encoders.
\STATE // \textit{Training phase.}
\FOR{$epoch=1,...,E$}
\STATE $\mathcal{B}\leftarrow$Randomly split $V$ into batches of size $B$.
\FOR{$b=(v_1',...,v_B')\in \mathcal{B}$}
\STATE $\mathbf{E}\leftarrow\text{MLP}(\{\mathbf{x}_{v_1'}, ...,\mathbf{x}_{v_B'}\})$
\STATE $\mathbf{E}_{\text{nei}}\leftarrow$Compute embeddings for neighbors\\  of $b$ using MLP encoder.
\STATE $\mathbf{A}_\text{nei}\leftarrow$Obtain neighborhood subgraphs of $b$.
\STATE $\bar{\mathbf{E}}=\text{GNN}(\mathbf{E}\|\mathbf{E}_{\text{nei}}, \mathbf{A}_\text{nei})$
\STATE Calculate the predicted scores $\mathbf{s}, \bar{\mathbf{s}}$ using $\mathbf{E}, \bar{\mathbf{E}}$ via Equation (4).
\STATE Calculate the ranking loss $\mathcal{L}_{\text{rank}}, \bar{\mathcal{L}}_{\text{rank}}$ using $\mathbf{s}, \bar{\mathbf{s}}, \mathbf{h}$ via Equation (8).
\STATE Calculate the asymmetric alignment loss $\mathcal{L}_\text{align}$ using $\mathbf{E}, \bar{\mathbf{E}}$ via Equation (10).
\STATE $\mathcal{L}\leftarrow \mathcal{L}_{\text{rank}} + \bar{\mathcal{L}}_{\text{rank}} + \alpha \mathcal{L}_\text{align}$
\STATE Back-propagate $\mathcal{L}$ to update the parameters of MLP and GNN with learning rate lr. 
\ENDFOR
\ENDFOR
\STATE // \textit{Inference phase.}
\STATE $\mathbf{E}\leftarrow\text{MLP}(\mathbf{X})$
\STATE Calculate the predicted scores $\mathbf{s}$ using $\mathbf{E}$ via Equation (4).
\STATE \textbf{return} $\mathbf{s}$
\end{algorithmic}
\end{algorithm}

\appendix
\section{Appendix E: Implementation Details}
\subsection{Environment}
\ourmethod is implemented with the following libraries and their respective versions: Python~3.9.18, CUDA version~11.7, PyTorch~2.0.0, DGL~1.1.2, torch-cluster~1.6.3, torch-sparse~0.6.18. 

\begin{table*}[]
\renewcommand\thetable{A.3} 
\fontsize{9pt}{12pt}\selectfont
\begin{tabular}{l|l|c|c|c|c|c|c}
 \toprule
\multicolumn{1}{l|}{\multirow{2}{*}{Metric}} & \multicolumn{1}{c|}{\multirow{2}{*}{Method}} & \multicolumn{6}{c}{Dataset}  \\ \cline{3-8} 
\multicolumn{1}{l|}{}& \multicolumn{1}{c|}{}& \multicolumn{1}{c|}{Amazon} & \multicolumn{1}{c|}{Facebook} & \multicolumn{1}{c|}{Reddit} & \multicolumn{1}{c|}{YelpChi} & \multicolumn{1}{c|}{AmazonFull} & \multicolumn{1}{c}{YelpChiFull} \\ 

\midrule
\multirow{8}{*}{AUROC} & DOMINANT& 0.4845±0.0064 & 0.4372±0.0046 & 0.5588±0.0043 & 0.3993±0.0030 & 0.4496±0.0683 & OOM   \\
& CoLA& 0.4734±0.0044 & 0.8366±0.0268 & \textbf{0.6032±0.0057} & 0.4336±0.0177 & 0.2110±0.0036 & \underline{0.4911±0.0008} \\
& ANEMONE & 0.5757±0.0056 & 0.8385±0.0104 & 0.5853±0.0047 & 0.4432±0.0129 & 0.6056±0.0037 & 0.4601±0.0017 \\
& GRADATE & 0.5539±0.0063 & 0.8809±0.0100 & 0.5671±0.0066 & OOM   & OOM   & OOM   \\
& ADA-GAD & 0.5240±0.0008 & 0.0756±0.0007 & 0.5610±0.0002 & OOM   & OOM   & OOM   \\
& GADAM   & 0.6167±0.0133 & \underline{0.9539±0.0067} & 0.5809±0.0053 & 0.4177±0.0168 & 0.4457±0.0205 & 0.4797±0.0085 \\
& TAM & \underline{0.7126±0.0111} & 0.8895±0.0057 & 0.5748±0.0021 & \underline{0.5473±0.0022} & \underline{0.6442±0.0232} & OOM   \\
& \textbf{HUGE (ours)} & \textbf{0.8516±0.0029} & \textbf{0.9760±0.0008} & \underline{0.5906±0.0042} & \textbf{0.6013±0.0076} & \textbf{0.8892±0.0026} & \textbf{0.5767±0.0064} \\
\midrule
\multirow{8}{*}{AUPRC}& DOMINANT& 0.0589±0.0007 & 0.0194±0.0002 & 0.0371±0.0002 & 0.0385±0.0002 & 0.0574±0.0081 & OOM   \\
& CoLA& 0.0683±0.0007 & 0.2155±0.0417 & 0.0440±0.0015 & 0.0448±0.0018 & 0.0536±0.0013 & \underline{0.1419±0.0004} \\
& ANEMONE & 0.1054±0.0054 & 0.2320±0.0299 & 0.0417±0.0008 & 0.0473±0.0017 & \underline{0.2396±0.0133} & 0.1305±0.0007 \\
& GRADATE & 0.0892±0.0021 & 0.3560±0.0300 & 0.0389±0.0007 & OOM   & OOM   & OOM   \\
& ADA-GAD & 0.1108±0.0006 & 0.0122±0.0000 & 0.0382±0.0021 & OOM   & OOM   & OOM   \\
& GADAM   & 0.0857±0.0050 & \underline{0.3630±0.0144} & \underline{0.0465±0.0018} & 0.0423±0.0015 & 0.0566±0.0027 & 0.1351±0.0028 \\
& TAM & \underline{0.2915±0.0373} & 0.1937±0.0172 & 0.0438±0.0005 & \textbf{0.0780±0.0014} & 0.2188±0.0314 & OOM   \\
& \textbf{HUGE (ours)} & \textbf{0.6672±0.0077} & \textbf{0.3674±0.0091} & \textbf{0.0511±0.0022} & \underline{0.0708±0.0026} & \textbf{0.7668±0.0022} & \textbf{0.1869±0.0026} \\
\bottomrule
\end{tabular}

\caption{Complete results (mean ± std) of main experiment. The best and second-best results are in \textbf{bold} and \underline{underlined}, respectively. OOM indicates out-of-memory on a 24GB GPU.}
\label{table:fullmainexperiment}
\end{table*}

\begin{table*}[t!]
\renewcommand\thetable{A.4} 

\centering
\begin{tabular}{l | c c c c c c } 
 \toprule
 Measures & Amazon & Facebook & AmazonFull & YelpChiFull  \\
 \midrule
  \begin{tabular}{@{}c@{}}\textbf{\ourmethod} \\ \textbf{(w/ HALO)}\end{tabular}   & \textbf{0.8516±0.0029} & \textbf{0.9760±0.0008} & \textbf{0.8892±0.0026} & \textbf{0.5767±0.0064} \\

 \midrule
w/ Euc. Dist.\ & 0.7128±0.0061 & 0.9251±0.0074 & 0.8746±0.0026 & 0.5763±0.0064 \\
w/ Cos. Dist. & 0.7018±0.0052 & 0.9668±0.0024 & 0.8735±0.0026 & 0.5762±0.0063  \\
w/ AHR & 0.8171±0.0064 & 0.9438±0.0029 & 0.8622±0.0024 &  0.5765±0.0064\\
\midrule
w/o Alignment\ & 0.5614±0.0019 & 0.9487±0.0026 & 0.8746±0.0033 & 0.5295±0.0204 \\
w/o GNN & 0.5548±0.0024 & 0.9401±0.0022 & 0.7005±0.0136  & 0.5533±0.0088  \\
 \bottomrule
\end{tabular}
\caption{Complete results (mean±std) of ablation study of key designs in \ourmethod. 
}

\label{table:complete_ablation}
\end{table*}
\subsection{Hardware Configuration}
All the experiments are conducted on a Windows desktop equipped with an AMD Ryzen 5800X processor, 32 GB of RAM, and a single NVIDIA GeForce RTX4090 GPU with 24GB of VRAM. 
\subsection{Hyper Parameters}
For the backbone of \ourmethod, we use two layers of MLPs followed by one layer of GNN, with 128 hidden perceptrons. In addition, we summarize other hyper-parameters of \ourmethod in Table~\ref{table:parameters}. By default, we utilize a learning rate of 0.0005, epoch of 300, batch size of 8192 and set $\alpha$ to 0.5. For the large datasets Amazon-Full and YelpChi-Full, we reduce the batch size to 512 to avoid out-of-memory issue. we also adjust the training epochs to 10 and 5 respectively to accelerate training, while increase the $\alpha$ to place greater emphasis on neighbor information. We obtain the best combination of hyper-parameters via grid search. The details of grid search are summarized in Table~\ref{table:gridsearch}. For all experiments, we record the average performance and standard deviation over five runs, using random seeds $\{0, 1, 2, 3, 4\}$.

\appendix
\section{Appendix F: More Results}

In this section, we propose additional experimental results that are not included in the main paper due to the length constraints. The complete results of the main experiments and ablation study are included in Table~\ref{table:fullmainexperiment} and Table~\ref{table:complete_ablation}, respectively.

\end{document}